\documentclass[table]{article}
\usepackage[accepted]{icml2016}
\usepackage{times}
\usepackage{amsmath}
\usepackage{amsthm}
\usepackage{xcolor}
\usepackage{subfig}
\usepackage{array}
\usepackage[draft]{hyperref}
\usepackage{natbib}
\usepackage[pdftex]{graphicx}
\graphicspath{ {./graphs/}{./graphs/feature_selection/}}
\setcitestyle{square}

\icmltitlerunning{Interpretable Selection and Visualization of Features and Interactions Using Bayesian Forests}

\begin{document} 
	
\twocolumn[
\icmltitle{Interpretable Selection and Visualization of Features and Interactions Using Bayesian Forests}

\icmlauthor{Viktoriya Krakovna}{vkrakovna@fas.harvard.edu}
\icmladdress{Department of Statistics,
	Harvard University,
	Cambridge, MA 02138}
\icmlauthor{Jiong Du}{}
\icmladdress{Haitong Securities}
\icmlauthor{Jun S. Liu}{jliu@stat.harvard.edu}
\icmladdress{Department of Statistics,
	Harvard University,
	Cambridge, MA 02138}
\vskip 0.3in
]

\begin{abstract}
It is becoming increasingly important for machine learning methods to make predictions that are interpretable as well as accurate. In many practical applications, it is of interest which features and feature interactions are relevant to the prediction task. We present a novel method, Selective Bayesian Forest Classifier, that strikes a balance between predictive power and interpretability by simultaneously performing classification, feature selection, feature interaction detection and visualization. It builds parsimonious yet flexible models using tree-structured Bayesian networks, and samples an ensemble of such models using Markov chain Monte Carlo. We build in feature selection by dividing the trees into two groups according to their relevance to the outcome of interest. Our method performs competitively on classification and feature selection benchmarks in low and high dimensions, and includes a visualization tool that provides insight into relevant features and interactions. 
\end{abstract}

\section{Introduction}


Feature selection and classification are key objectives in machine learning that are usually tackled separately. However, performing classification on its own tends to produce black box solutions that are difficult to interpret, while performing feature selection alone can be difficult to justify without being validated by prediction. In addition to screening for relevant features, it is also useful to detect interactions between them. In many decision support systems, e.g. in medical diagnostics, the users care about which features and interactions contributed to a particular decision made by the system. Selective Bayesian Forest Classifier (SBFC) combines predictive power and interpretability by performing classification, feature selection, and feature interaction detection at the same time. Our method also provides a visual representation of the relevance of different features and feature interactions to the outcome of interest. 
 
\begin{figure}
	\centering
	\includegraphics[scale=0.3]{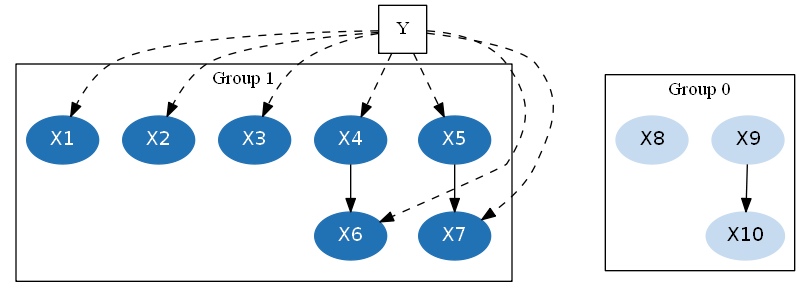}
	\caption{Example of a SBFC graph}\label{fig:graph_sbfc}
	\vskip -2em
\end{figure}

The main idea of SBFC is to construct an ensemble of Bayesian networks \citep{Pearl}, each constrained to a forest of trees divided into signal and noise groups based on their relationship with the class label $Y$ (see Figure \ref{fig:graph_sbfc} for an example). The nodes and edges in Group 1 represent relevant features and interactions. 
Such models are easy to sample using Markov chain Monte Carlo (MCMC). We combine their predictions using Bayesian model averaging, and aggregate their feature and interaction selection.

We show that SBFC performs competitively with state-of-the-art methods on 25 low-dimensional and 6 high-dimensional benchmark data sets. By adding noise features to a synthetic data set, we compare feature selection and interaction detection performance as the signal to noise ratio decreases (Figure \ref{fig:corral_graphs}). We use a high-dimensional data set from the NIPS 2003 feature selection challenge to demonstrate SBFC's superior performance on a difficult feature selection task (Figure \ref{fig:madelon_graphs}), and illustrate the visualization tool on a heart disease data set with meaningful features (Figure \ref{fig:heart_graphs}). SBFC is a good choice of algorithm for applications where interpretability matters along with predictive power (an R package is available at \url{github.org/vkrakovna/sbfc}).



\section{Related Work} \label{sec:related}

Tree structures are frequently used in computer science and statistics, because they provide adequate flexibility to model complex structures, yet are constrained enough to facilitate computation. SBFC was inspired by tree-based methods such as Tree-Augmented Na{\"i}ve Bayes (TAN) \citep{FriedmanBN} and Averaged One-Dependence Estimators (AODE) \citep{Webb}. 
TAN finds the optimal tree on all the features using the minimum spanning tree algorithm, with the class label $Y$ as a second parent for all the features. 
While the search for the best unrestricted Bayesian network is usually an intractable task \citep{Heckerman}, the computational complexity of TAN is only $O(d^2 n)$, where $d$ is the number of features and $n$ is the sample size \citep{Chow}. 
AODE constrains the model structure to a tree where all the features are children of the root feature, with $Y$ as a second parent, and uses model averaging over model with all possible root features.
These methods put all the features into a single tree, which can be difficult to interpret, especially for high-dimensional data sets. 
We extend on TAN and AODE by building forests instead of single-tree graphs, and introducing selection of relevant features and interactions. 



Feature selection is often used as a preprocessing step for classification algorithms. Wrapper methods \citep{Kohavi} select a subset of features tailored for a specific classifier, treating it as a black box. Variable Selection for Clustering and Classification (VSCC) \citep{Andrews} searches for a feature subset that simultaneously minimizes the within-class variance and maximizes the between-class variance, and remains efficient in high dimensions. Categorical Adaptive Tube Covariate Hunting (CATCH) \citep{Tang} selects features based on a nonparametric measure of the relational strength between the feature and the class label. 

Our approach, however, is to integrate feature selection into the classification algorithm itself, allowing it to influence the models built for classification. A classical example is Lasso \citep{TibshiraniLasso}, which performs feature selection using $L_1$ regularization. Some decision tree classifiers, like Random Forest \citep{BreimanRF} and BART \citep{Chipman}, provide importance measures for features and the option to drop the least significant features. 
In many applications, it is also key to identify relevant feature interactions, such as epistatic effects in genetics. Interaction detection methods for gene association models include Graphical Gaussian models \citep{Andrei} and Bayesian Epistasis Association Mapping (BEAM) \citep{ZhangBEAM}. BEAM introduces a latent indicator that partitions the features into several groups based on their relationship with the class label. One of the groups in BEAM is designed to capture relevant feature interactions, but is only able to tractably model a small number of them. SBFC extends this framework, using tree structures to represent an unlimited number of relevant feature interactions. 


\section{Selective Bayesian Forest Classifier (SBFC)} \label{sec:SBFC}


\subsection{Model}
Given $n$ observations with class label $Y$ and $d$
discrete features $X_{j}$, $j=1,\dots,d$, we divide the features into two groups based on their relation to
$Y$:
\begin{description}
\item [{Group~0~(noise):}] features that are unrelated to $Y$ 
\item [{Group~1~(signal):}] features that are related to $Y$ 
\end{description}

We further partition each group into non-overlapping subgroups
mutually independent of each other conditional on $Y$. 
For each subgroup, we infer a tree structure describing the dependence relationships
between the features (many subgroups will consist of one node and thus have a trivial dependence structure). Note that we model the structure in the noise group as well as the signal group, since an independence assumption for the noise features could result in correlated noise features being misclassified as signal features. 

The overall dependence structure is thus modeled as a forest of trees, representing conditional dependencies between the features (no causal relationships are inferred).
The class label $Y$ is a parent of every feature in Group 1 (edges to $Y$ are omitted in subsequent figures). We will refer to the combination of a group partition and a forest structure as a graph.
 
The prior consists of a penalty on the number of edges between features in each group and a penalty on the number of signal nodes (i.e., edges between features and $Y$)
\[ P(G) \propto d^{- 4 (E_0(G) + E_1(G) /v) - D_1(G)/v} \]
where $D_i(G)$ is the number of nodes and $E_i(G)$
is the number of edges in Group $i$ of graph $G$, while $v$ is a constant equal to the number of classes. 

The prior scales with $d$, the number of features, to penalize very large, hard-to-interpret trees in high dimensional cases. The terms corresponding to the signal group are divided by $v$, the number of possible classes, to avoid penalizing large trees in the signal group more than in the noise group by default. The coefficients in the prior were found in practice to provide good classification and feature selection performance (there is a relatively wide range of coefficients that produce similar results). 

Given the training data $X_{(n\times d)}$ (with columns $\boldsymbol{X}_{j}$,
$j=1,\dots,d$) and $\boldsymbol{y}_{(n\times1)}$, we break down
the graph likelihood according to the tree structure:
\begin{align*}
P(X,\boldsymbol{y}|G) & =P(\boldsymbol{y}|G)P(X|\boldsymbol{y},G)\\
 & =P(\boldsymbol{y})\prod_{j=1}^{d}P(\boldsymbol{X}_{j}|\boldsymbol{\Lambda}_{j})
\end{align*}
Here, $\Lambda_{j}$ is the set of parents of $X_{j}$ in
graph $G$. This set includes the parent $X_{p_{j}}$ of
$X_{j}$ unless $X_{j}$ is a root, and $Y$ if $X_{j}$ is in Group
1, as shown in Table \ref{tab:parent_set}. 
We assume that the distributions of the class label $Y$ and the graph structure $G$ are independent a priori.

\begin{table}
\caption{Parent sets for each feature type\label{tab:parent_set}}
\centering
\begin{tabular}{|c|c|}
\hline 
Type of feature $X_{j}$ & Parent set $\Lambda_{j}$ \\
\hline
Group 0 root & $\emptyset$ \\
Group 0 non-root & $\{X_{p_{j}}\}$ \\
Group 1 root & $\{Y\}$ \\
Group 1 non-root & $\{Y,X_{p_{j}}\}$ \\
\hline 
\end{tabular}
\end{table}

Let $v_{j}$ and $w_{j}$ be the number of possible values for $X_{j}$
and $\Lambda_{j}$ respectively. Then our hierarchical model for $X_{j}$
is 
\begin{align*}
[X_{j}  |\Lambda_{j}=\Lambda_{jl}, \boldsymbol{\Theta}_{jl} = \boldsymbol{\theta}_{jl}] &\sim\mbox{Mult}(\boldsymbol{\theta}_{jl}), \; l = 1,\dots, w_j\\
\boldsymbol{\Theta}_{jl} & \sim\mbox{Dirichlet}\left(\frac{\alpha}{w_{j}v_{j}}\boldsymbol{1}_{v_{j}}\right)
\end{align*}

Each conditional Multinomial model has a different parameter vector
$\boldsymbol{\Theta}_{jl}$. We consider the Dirichlet hyperparameters
to represent ``pseudo-counts'' in each conditional model \citep{FriedmanBN}. 
Let $n_{jkl}$ be the number of observations
in the training data with $X_{j}=x_{jk}$ and $\Lambda_{j}=\Lambda_{jl}$,
and $n_{jl}=\sum_{k=1}^{v_{j}}n_{jkl}$. Then 
\[
P(\boldsymbol{X}_{j}|\boldsymbol{\Lambda}_{j}, \boldsymbol{\Theta}_{j1},\dots, \boldsymbol{\Theta}_{jw_j}) =\prod_{l=1}^{w_{j}}\prod_{k=1}^{v_{j}}\theta_{jkl}^{n_{jkl}}
\]

We then integrate out the nuisance parameters $\boldsymbol{\Theta}_{jl}$, $l=1,\dots,w_j$.
The resulting likelihood depends only on the hyperparameter $\alpha$ and the
counts of observations for each combination of values of $X_{j}$ and $\Lambda_{j}$.
\[
P(\boldsymbol{X}_{j}|\boldsymbol{\Lambda}_{j})=\prod_{l=1}^{w_{j}}\frac{\Gamma\left(\frac{\alpha}{w_{j}}\right)}{\Gamma\left(\frac{\alpha}{w_{j}}+n_{jl}\right)}\prod_{k=1}^{v_{j}}\frac{\Gamma\left(\frac{\alpha}{w_{j}v_{j}}+n_{jkl}\right)}{\Gamma\left(\frac{\alpha}{w_{j}v_{j}}\right)}
\]

This is the Bayesian Dirichlet score, which satisfies likelihood
equivalence \citep{Heckerman}. Namely, reparametrizations of the model that do not affect the conditional independence relationships between
the features, for example by pivoting a tree to a different root, do not change the likelihood.

\begin{figure}[h]
\centering
\subfloat[Switch Trees: switch tree $\{X_5,X_7\}$ to Group 0, switch tree $\{X_8\}$ to Group 1]{\includegraphics[scale=0.3]{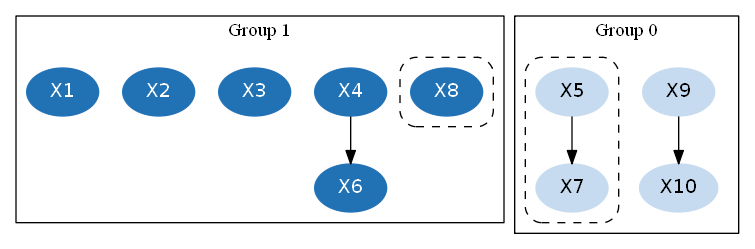}\label{subfig:graph_switch}}\\
\subfloat[Reassign Subtree: reassign node $X_6$ to be a child of node $X_8$]{\includegraphics[scale=0.3]{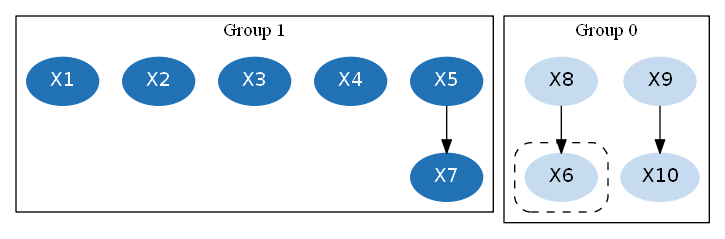}\label{subfig:graph_reassign}}\\
\subfloat[Pivot Trees: nodes $X_6$ and $X_{10}$ become tree roots]{\includegraphics[scale=0.3]{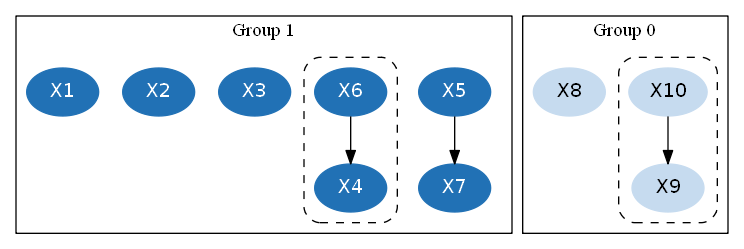}\label{subfig:graph_pivot}}\\

\caption{Example MCMC updates applied to the graph in Figure \ref{fig:graph_sbfc}}\label{fig:graph_updates}
\end{figure}

\subsection{MCMC Updates}
\begin{description}
\item [{Switch Trees:}] Randomly choose trees $T_1,\dots,T_k$ without replacement (we use $k=10$, and propose switching each tree to the opposite group one by one (see Figure \ref{subfig:graph_switch}). This is a repeated Metropolis update.


\item [{Reassign Subtree:}]  Randomly choose a node $X_{j}$, detach the subtree rooted at this node and choose a different parent node for this subtree (see Figure \ref{subfig:graph_reassign}). This is a Gibbs update, so it is always accepted.


We consider the set of nodes $X_{j'}$ that are not descendants of $X_{j}$ as candidate parent nodes (to avoid creating a cycle), with corresponding graphs $G_{j'}$. We also consider a ``null parent" option for each group, where $X_j$ becomes a root in that group, with corresponding graph $\tilde G_i$ for group $i$. Choose a graph $G^*$ from this set according to the conditional posterior distribution $\pi(G^*)$ (conditioning on the parents of all the nodes except $X_j$, and on the group membership of all the nodes outside the subtree). The subtree joins the group of its new parent. 


As a special case, this results in a tree merge if $X_{j}$
was a root node, or a tree split if $X_{j}$ becomes a root (i.e.
the new parent is null). Note that the new parent can be the original parent, in which case the graph does not change. 

\item[{Pivot Trees:}] Pivot all the trees by randomly choosing a new root
for each tree (see Figure \ref{subfig:graph_pivot}). By likelihood equivalence, this update is always accepted.

For computational efficiency, in practice we don't pivot all the trees at each
iteration. Instead, we just pivot the tree containing the chosen node $X_{j}$
within each Reassign Subtree move, since this is the only time
the parametrization of a tree matters. This implementation produces
an equivalent sampling mechanism. 

\end{description}

\begin{table}[h]
\caption{ Data set properties \citep{FriedmanBN}} \label{tab:dataset_summary}
\centering
\small
\setlength{\tabcolsep}{.4em}
\begin{tabular}{|>{\texttt\bgroup}l<{\egroup}|c|c|cc|}\hline
  \normalfont{Data set} &\#Features &\#Classes &\multicolumn{2}{c|}{\#Instances}\\
              &              &           &Train &Test\\\hline
australian    &14 &2  &690   &CV-5\\
breast        &10 &2  &683   &CV-5\\
chess         &36 &2  &2130  &1066\\
cleve         &13 &2  &296   &CV-5\\
corral        &6  &2  &128   &CV-5\\
crx           &15 &2  &653   &CV-5\\
diabetes      &8  &2  &768   &CV-5\\
flare         &10 &2  &1066  &CV-5\\
german        &20 &2  &1000  &CV-5\\
glass 	& 9 & 6 & 214 & CV-5 \\
glass2        &9  &2  &163   &CV-5\\
heart         &13 &2  &270   &CV-5\\
hepatitis     &19 &2  &80    &CV-5\\
iris 		& 4 & 3 & 150 & CV-5 \\
letter 	& 16 & 26 & 15000 & 5000 \\
lymphography & 18 & 4 & 148 & CV-5 \\
mofn-3-7-10   &10 &2  &300   &1024\\
pima          &8  &2  &768   &CV-5\\
satimage & 36& 6& 4435 & 2000\\
segment & 19 & 7 & 1540 & 770\\
shuttle-small & 9& 6& 3866 & 1934\\
soybean-large &  35 & 19 & 562 & CV-5\\
vehicle 	&18 & 4 & 846 & CV-5\\
vote          &16 &2  &435   &CV-5\\
waveform-21 & 21 &3 &300 &4700\\
\hline\hline
ad & 1558 & 2 & 2276 & 988 \\
arcene & 10000 &  2 & 100 & 100 \\
arcene-cv & 10000 & 2 & 200 & CV-5\\
gisette & 5000 & 2 & 6000 & 1000 \\
isolet & 617 & 26 & 6238 & 1559 \\
madelon & 500 & 2 & 2000 & 600\\
microsoft & 294 & 2 & 32711 & 5000 \\
\hline
\end{tabular}
\end{table}

\subsection{Classification Using Bayesian Model Averaging}

Graphs are sampled from the posterior distribution using the MCMC algorithm. We apply Bayesian model averaging \citep{Hoeting} rather than using the posterior mode for classification. For each possible class, we average the probabilities over a thinned subset of the sampled graph structures, and then choose the class label with the highest average probability. Given a test data point $\boldsymbol{x}^{\mbox{test}}$, we find
\begin{align*}
&P(Y= y |\boldsymbol{X}= \boldsymbol{x}^{\mbox{test}}, X, \boldsymbol{y})\\
\propto&\sum_{i=1}^SP(Y=y|\boldsymbol{X}= \boldsymbol{x}^{\mbox{test}},G_i) P(G_i | X, \boldsymbol{y})
\end{align*}
where $S$ is the number of graphs sampled by MCMC (after thinning by a factor of 50). We use training data counts to compute the posterior probability of the class label given each sampled graph $G_i$.

\begin{figure*}[t!]
\centering
\includegraphics[scale=.75]{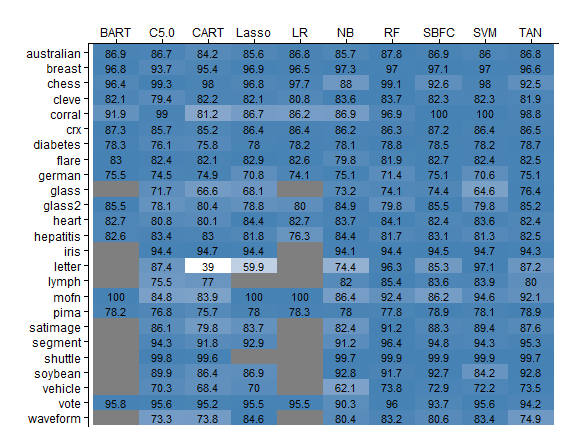}\\
\vspace{-2em}
\includegraphics[scale=.75]{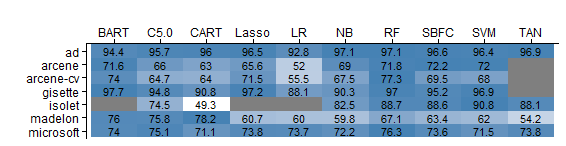}
\caption{Classification accuracy on low- and high-dimensional data sets, showing
average accuracy over 5 runs for each method, with the top half of the methods in bold for each data set. Note that some of the classifiers could not handle multiclass data sets, and TAN timed out on the highest-dimensional data sets. SBFC performs competitively with SVM, TAN and some decision tree methods (BART and RF), and generally outperforms the others.}\label{fig:classification}
\end{figure*}

\begin{figure*}[t!]
\centering
\subfloat[A sampled graph for \texttt{heart} data set]{\includegraphics[scale=.6]{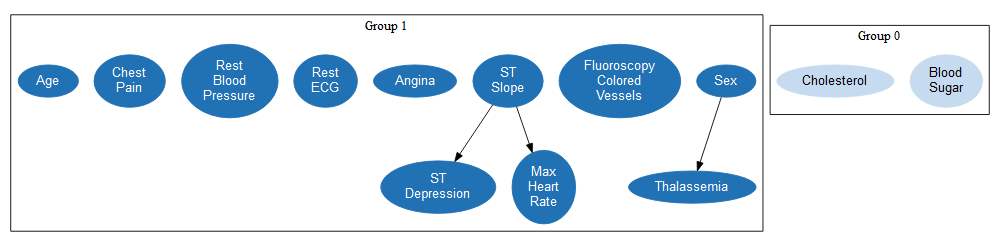}\label{subfig:heart_samp}}\\
\subfloat[Average graph for \texttt{heart} data set]{\includegraphics[scale=.6]{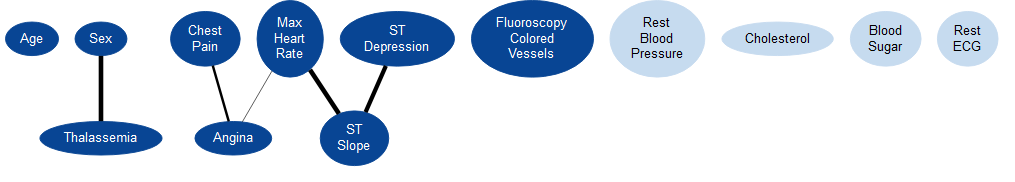}\label{subfig:heart_average}}\\
\subfloat[Feature selection comparison for \texttt{heart} data set]{\includegraphics[scale=.75]{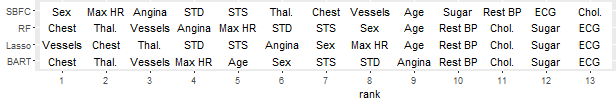}\label{subfig:heart_rank}}
\caption{The sampled graph in Figure \ref{subfig:heart_samp} and the average graph in Figure \ref{subfig:heart_average} show feature and interaction selection for the \texttt{heart} data set with features of medical significance. The dark-shaded features in the average graph are the most relevant for predicting heart disease. There are several groups of relevant interacting features: (Sex, Thalassemia), (Chest Pain, Angina), and (Max Heart Rate, ST Slope, ST Depression). The features in each group jointly affect the presence of heart disease. Figure \ref{subfig:heart_rank} compares feature rankings with other methods, showing that all the methods agree on the top 9 features, but SBFC disagrees with the other methods on the top 3 features.}\label{fig:heart_graphs}
\end{figure*}

\begin{figure*}[p]
\centering
\subfloat[A sampled graph for the original \texttt{corral} data set with 6 features]{\includegraphics[scale=0.4]{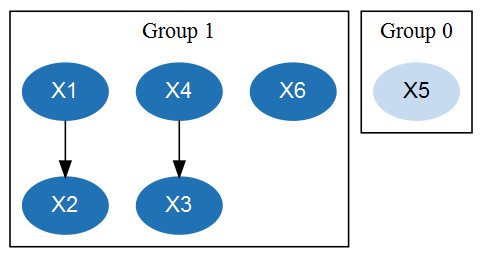}\label{subfig:corral6_samp}}
\hspace{2em}
\subfloat[A sampled graph for the augmented \texttt{corral} data set with 100 features]{\includegraphics[scale=0.8]{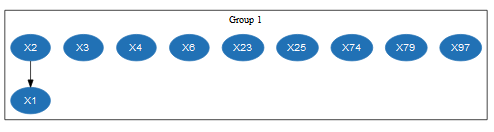}\label{subfig:corral100_samp}}\\
\subfloat[Average graph for the original \texttt{corral} data set with 6 features]{\includegraphics[scale=0.5]{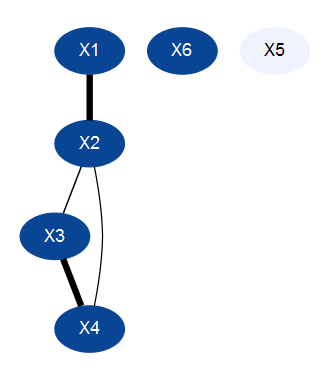}\label{subfig:corral6_average}}
\hspace{10em}
\subfloat[Average graph for the augmented \texttt{corral} data set with 100 features]{\includegraphics[scale=0.5]{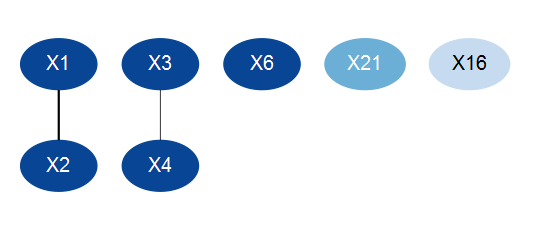}\label{subfig:corral100_average}}\\
\subfloat[Feature selection comparison for the original \texttt{corral} data set with 6 features]{\includegraphics[scale=.8]{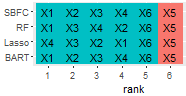}\label{subfig:corral6_rank}}
\hspace{8em}
\subfloat[Feature selection comparison for the augmented \texttt{corral} data set with 100 features]{\includegraphics[scale=.8]{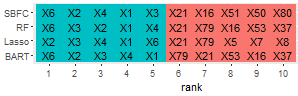}\label{subfig:corral100_rank}}\\
\caption{In the synthetic data set \texttt{corral}, the true feature structure is known: the relevant features are $\{X_1,X_2,X_3,X_4,X_6\}$, and the most relevant edges are $\{X_1, X_2\}$, $\{X_3, X_4\}$, while the other edges between the first 4 features are less relevant, and any edges with $X_5$ or $X_6$ are not relevant. 
The sampled graph in Figure \ref{subfig:corral6_samp} and the average graph in Figure \ref{subfig:corral6_average} show that SBFC recovers the true correlation structure between the features, with the most relevant edges appearing the most frequently (as indicated by thickness). We generate extra noise features for this data set by choosing an existing feature at random and shuffling the rows, making it uncorrelated with the other features. The sampled graph in Figure \ref{subfig:corral100_samp} and the average graph in Figure \ref{subfig:corral100_average} show that SBFC recovers the relevant features and some relevant interactions when the amount of noise increases. Figures \ref{subfig:corral6_rank} and \ref{subfig:corral100_rank} show that all the methods consistently rank the 5 relevant features (colored blue) above the rest (colored red).}
\label{fig:corral_graphs}
\end{figure*}

\begin{figure*}[p]
\centering
\subfloat[A sampled graph for \texttt{madelon} data set]{\includegraphics[scale=.8]{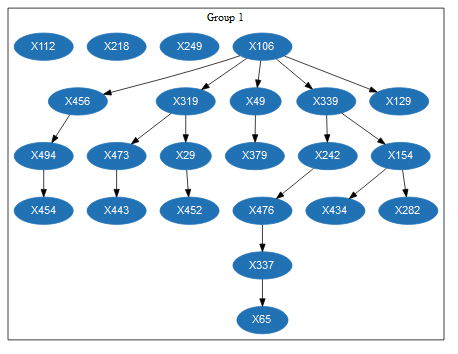}\label{subfig:madelon_samp}}\\
\subfloat[Average graph for \texttt{madelon} data set]{\includegraphics[scale=1]{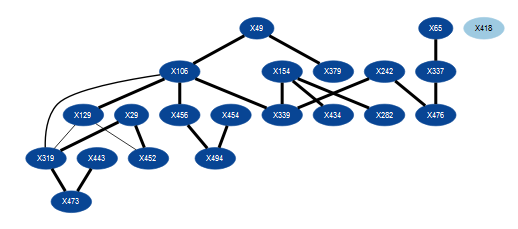}\label{subfig:madelon_average}}\\
\subfloat[Feature selection comparison for \texttt{madelon} data set]{\includegraphics[scale=.75]{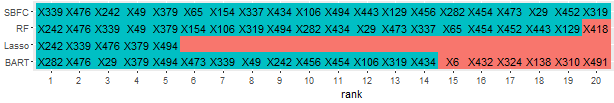}\label{subfig:madelon_rank}}
\caption{Feature and edge selection for the synthetic \texttt{madelon} data set, used in the 2003 NIPS feature selection challenge. This data set, with 20 relevant features and 480 noise features, was artificially constructed to illustrate the difficulty of selecting a feature set when no feature is informative by itself, and all the features are correlated with each other \citep{GuyonNIPS}. SBFC reliably selects the correct set of 20 relevant features \citep{Guyon}, as shown in Figure \ref{subfig:madelon_rank}, and appropriately puts them in a single connected component, shown in dark blue in the average graph in Figure \ref{subfig:madelon_average}. As shown in Figure \ref{subfig:madelon_rank}, none of the other methods correctly identify the set of 20 relevant features (colored blue), though Random Forest comes close with 19 out of 20 correct. Our classification performance on this data set is not as good as that of BART or RF, likely because SBFC constrains these highly correlated features to form a tree structured Bayesian network, while a decision tree structure allows a feature to appear more than once.}
\label{fig:madelon_graphs}
\end{figure*}

\begin{table}
\centering
\caption{SBFC runtime on high-dimensional data sets in minutes}
\label{tab:runtime}
\begin{tabular}{|>{\texttt\bgroup}l<{\egroup}|c|}\hline
\normalfont{Data set} & Runtime (min)\\\hline
ad & 5\\
arcene & 60\\
arcene-cv & 65\\
gisette & 134\\
isolet & 23\\
madelon & 1\\
microsoft & 2\\
\hline
\end{tabular}
\end{table}

\section{Experiments} \label{sec:testing}

We compare our classification performance with the following methods.
\begin{description}
\item[\rm BART:] Bayesian Additive Regression Trees, R package \texttt{BayesTree} \citep{Chipman},
\item[\rm C5.0:] R package \texttt{C50} \citep{Quinlan}, 
\item[\rm CART:] Classification and Regression Trees, R package \texttt{tree} \citep{BreimanCART}, 
\item[\rm Lasso:] R package \texttt{glmnet} \citep{FriedmanGLM},
\item[\rm LR:] logistic regression, 
\item[\rm NB:] Na{\"i}ve Bayes, R package \texttt{e1071} \citep{Duda}
\item[\rm RF:] Random Forest, R package \texttt{ranger} \citep{BreimanRF}, 
\item[\rm SVM:] Support Vector Machines, R package \texttt{e1071} \citep{Evgeniou}, 
\item[\rm TAN:] Tree-Augmented Na{\"i}ve Bayes, R package \texttt{bnlearn} \citep{FriedmanBN}.
\end{description}

We use 25 small benchmark data sets used by \citet{FriedmanBN} and 6 high-dimensional data sets \citep{GuyonNIPS}, all from the UCI repository \citep{Lichman}, described in Table \ref{tab:dataset_summary}. We split the large data sets into a training set and a test set, and use 5-fold cross validation for the smaller data sets (we try both approaches for the high-dimensional \texttt{arcene} data set). We remove the instances with missing values, and discretize continuous features, using Minimum Description Length Partitioning \citep{Fayyad} for the small data sets and binary binning \citep{Dougherty} for the large ones. For a data set with $d$ features, we run SBFC for $\max(10000, 10 d)$ iterations, which has empirically been sufficient for stabilization. Figure \ref{fig:classification} compares SBFC's classification performance to the other methods.


We evaluate SBFC's feature selection and interaction detection performance on the data sets \texttt{heart}, \texttt{corral}, and \texttt{madelon}, in Figures \ref{fig:heart_graphs}, \ref{fig:corral_graphs}, and \ref{fig:madelon_graphs} respectively. We compare SBFC's feature selection performance to Lasso, as well as RF's \texttt{importance} metric and BART's \texttt{varcount} metric, which rank features by their influence on classification, in Figures \ref{subfig:heart_rank}, \ref{subfig:corral6_rank}, \ref{subfig:corral100_rank}, and \ref{subfig:madelon_rank}. We illustrate the structures learned by SBFC on these data sets using sampled graphs, shown in Figures \ref{subfig:heart_samp}, \ref{subfig:corral6_samp}, \ref{subfig:corral100_samp}, and \ref{subfig:madelon_samp}, and average graphs over all the MCMC samples, shown in Figures \ref{subfig:heart_average}, \ref{subfig:corral6_average}, \ref{subfig:corral100_average}, and \ref{subfig:madelon_average}. 

In the average graphs, the nodes are color-coded according to relevance, based on the proportion of sampled graphs where the corresponding feature appeared in Group 1 (dark-shaded nodes appear more often). Edge thickness also corresponds to relevance, based on the proportion of samples where the corresponding feature interaction appeared. To avoid clutter, only edges that appear in at least 10\% of the sampled graphs are shown, and nodes that appear in Group 0 more than 80\% of the time are omitted for high-dimensional data sets. Average graphs are undirected and do not necessarily have a tree structure. They provide an interpretable visual summary of the relevant features and feature interactions. 


As shown in Table \ref{tab:runtime}, the runtime of SBFC scales approximately as $ d \cdot n \cdot 2 \cdot 10^{-4}$ seconds (on an AMD Opteron 6300-series processor), so it takes somewhat longer to run than many of the other methods on high-dimensional data sets. 
SBFC's memory usage scales quadratically with $d$. 

\section{Conclusion}\label{sec:discussion}

Selective Bayesian Forest Classifier is an integrated tool for supervised classification, feature selection, interaction detection and visualization. It splits the features into signal and noise groups according to their relationship with the class label, and uses tree structures to model interactions among both signal and noise features. The forest dependence structure gives SBFC modeling flexibility and competitive classification performance, and it maintains good feature and interaction selection performance as the signal to noise ratio decreases. Useful directions for future work include extending SBFC to a semi-supervised learning method, and improving runtime and memory performance. 

\clearpage
\bibliography{sbfc_bibliography}
\bibliographystyle{icml2016}

\end{document}